\documentclass[a4paper, 10pt, conference]{ieeeconf}      

\IEEEoverridecommandlockouts                              
\usepackage{graphics} 
\usepackage{epsfig} 
\usepackage{mathptmx} 
\usepackage{times} 
\usepackage{amsmath} 
\usepackage{amssymb}  

\usepackage{algorithm}
\usepackage{algpseudocode}

\title{\LARGE \bf
Effectiveness of greedily collecting items in open world games
}

\author{Andrej Gajduk
\thanks{*This work was not supported by any organization}
\thanks{Source code available at https://github.com/gajduk/greedy-tsp}%
}

\begin{document}

\maketitle
\thispagestyle{empty}
\pagestyle{empty}

\begin{abstract}
Since \textit{Pokemon Go} sent millions on the quest of collecting virtual monsters, an important question has been on the minds of many people:
Is going after the closest item first a time-and-cost-effective way to play? Here, we show that this is in fact a good strategy which performs on average only $\textbf{7\%}$ worse than the best possible solution in terms of the total distance traveled to gather all the items. Even when accounting for errors due to the inability of people to accurately measure distances by eye, the performance only goes down to $\textbf{16\%}$ of the optimal solution.
\end{abstract}

\section{INTRODUCTION}

The traveling salesman problem (TSP) is a well known NP-hard problem. It asks the following question: \textit{How can a traveling salesman visit all of $N$ cities at least once and return to his starting location in a most efficient manner?} Efficient in this context can refer to distance, time or money. Because of its importance in real world scenarios such as delivery networks and supply chains it has been extensively studied and many optimization algorithms have been proposed in the literature~\cite{tsp_dynamic,tsp_genetic,tsp_natural}. However, humans have repeatedly been shown to perform better than computer algorithms on specific instances of the problem~\cite{humanz,humanz2}.

Another emerging scenario in which TSP is relevant are open world computer games. These games, sometimes also called free roam, allow players to freely explore a vast virtual world, completing missions and gathering collectibles in their preferred order. Some famous games that fall into this category are \textit{Grand Theft Auto}, \textit{World of Warcraft}, \textit{Assasins Creed} and \textit{Far Cry} featuring modern-urban, fantasy, historic and pacific island settings, respectively. Most recently, \textit{Pokemon GO} used augmented reality to make the whole world a setting for an open world game. All these games feature a range of items, hidden notes, virtual currency stashes or pokemons scattered throughout the virtual world. One of the objectives for the players is to collect as many of these collectibles as possible, especially if they are looking to earn the \textit{Completionist} achievement.

In order to investigate how players of open world games can effectively gather collectibles we focus our attention on a variation of the TSP, where the player must gather all the collectibles starting from a given position, but returning to the starting position is not required. Furthermore, the collectibles are located on a 2D map, and moving between any pair of collectibles is unobstructed as in most games. The optimization parameter is the total distance traveled in order to to collect all the collectibles.

Our goal is to see how a simple greedy algorithm compares to the optimal solution in realistic scenarios. The algorithm is outlined in Section~\ref{algorithm} and our finding related to its performance in Section~\ref{results}. Section~\ref{conclusion} concludes this manuscript.

\begin{figure}[b]
\centering
\includegraphics[scale=.25]{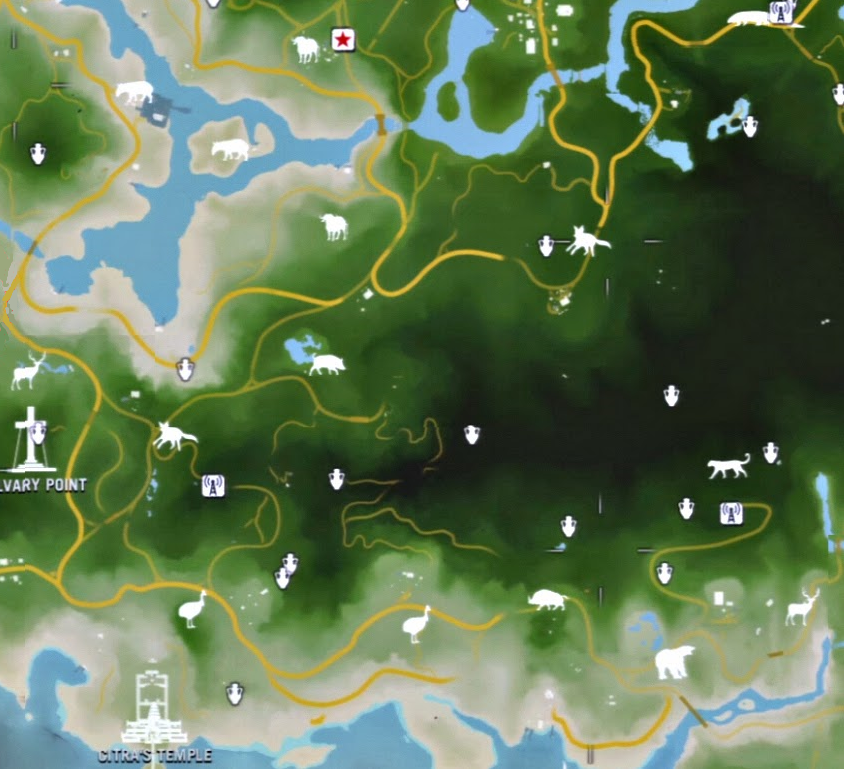}
\caption{An excerpt from the map from Far Cry 3. Pitchers denote the locations of collectibles}
\label{fig:map}
\end{figure}

\section{Algorithm}
\label{algorithm}

The algorithm of interest is based on the following gaming strategy: \textit{Always first gather the item which is closest to your current position}. This strategy is what a player would generally use because it is straightforward and does not require a lot of planning.

\begin{figure*}[t]
\centering
\includegraphics[width=\textwidth]{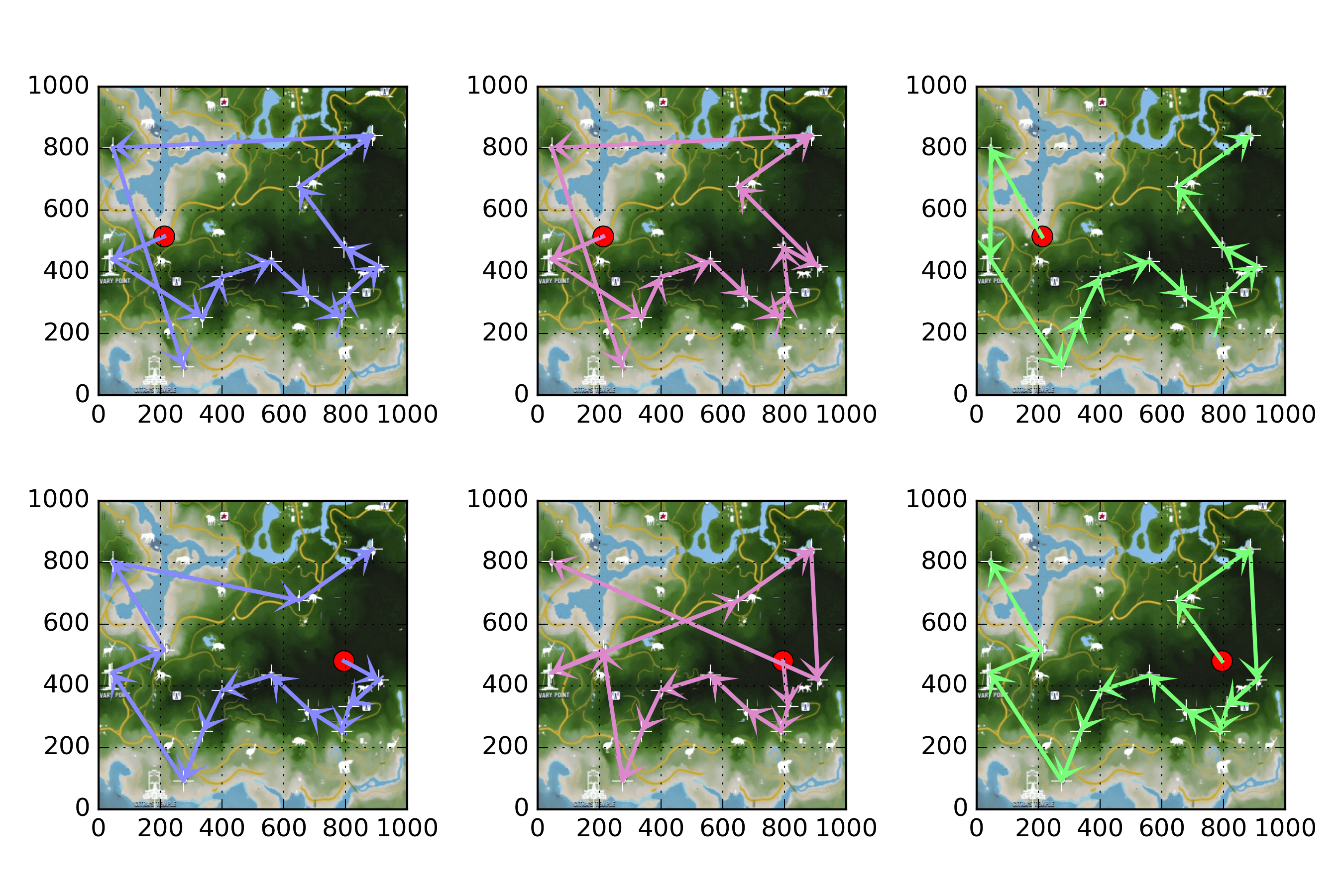}
\caption{Rows are different starting positions (red circle). Columns are different algorithms, from left to right: greedy, greedy with error ($\sigma=0.4$) and optimal. Total distances are for the first row $3601-3740-2764$; second row $2961-3923-2883$. The performance of the greedy algorithm depends on the choice of starting position with the difference being much larger for the first row ($30.2\%$ longer distance compared to the optimal) as compared to the second row ($2.4\%$). Furthermore the greedy algorithm with error clearly under-performs when compared to the greedy algorithm.}
\label{fig:starting_location_dependence}
\end{figure*}

To find the optimal solution an algorithm based on exhaustive search was used. It considers all possible permutations of the collectibles, and is thus guaranteed to find the optimal solution. Because the number of permutations is proportional to the factorial of the number of collectibles $N!$ this algorithm works only when $N \le 13$, which subsequently affects our analysis. The \textbf{greedy} algorithm works as follows:

\begin{algorithm}[!ht]
  \caption{Greedy algorithm}\label{greedy}
  \begin{algorithmic}[1]
    \Procedure{Greedy}{$start\_pos,C$}
    \State $current\_pos\gets start\_pos$
    \State $V\gets \lbrace \rbrace$
    \While{$|V| < |C|$}
      \State $next\_pos\gets \min_c dist(current\_pos,c), \forall c \in C \setminus V$
      \State $V\gets V\cup next\_pos$
      \State $current\_pos\gets next\_pos$
    \EndWhile
    \EndProcedure
  \end{algorithmic}
\end{algorithm}

where $C$ is a set of the positions of all the collectibles, and $dist$ is a function which determines the Euclidean distance between two positions in a 2D plane.

While a player is immersed in the game however, it might not be possible to measure the distance between two collectibles exactly, so the playes has to guestimate it, which may result in errors. To capture this, we propose the \textbf{greedy with error} algorithm which includes the following modification to the line 5 in the greedy algorithm
$$
next\_pos\gets \left[ \min_c dist(current\_pos,c) + \zeta \right], \forall c \in C \setminus V
$$

\begin{figure}[!b]
\centering
\includegraphics[scale=.45]{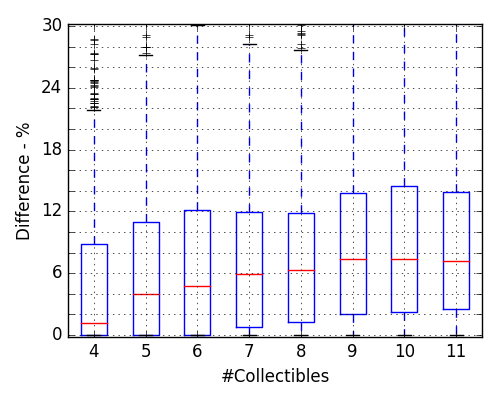}
\caption{Average performance of the greedy algorithm compared to the optimal solution, for different number of collectibles.}
\label{fig:box_plots}
\end{figure}

where $\zeta$ is an error term from a normal distribution $\mathcal{N}(1,\sigma)$ with cutoffs at $0.7$ and $1.3$ to prevent unreasonable errors.

\section{Results}
\label{results}

For illustration,  an excerpt  from the Far Cry 3 world will be used, where collectibles are marked by pitchers~(Fig.~\ref{fig:map}).

Not surprisingly,  the performance of the greedy algorithm depends on the starting location as shown in Figure~\ref{fig:starting_location_dependence}. In the first case, the route chosen by the greedy algorithm is suboptimal and results in a total distance of $3601$ (Figa.~\ref{fig:starting_location_dependence}-upper left), opposed to the optimal distance of $2764$ (Fig.~\ref{fig:starting_location_dependence}-upper right), which is a difference of $30.2\%$. In the second case a different starting location is chosen which results in the greedy algorithm finding a solution that is much close to the optimal one ($2961-2883$ or a difference of $2.4\%$), as indicated by the similar routes in the second row of Figure\ref{fig:starting_location_dependence}.

Although, this example is intriguing a detailed analysis is needed to quantify the performance of the greedy algorithm. To this end the following setup was used: $N$ collectibles were randomly positioned on a 2D plane (1000x1000 in size), assuming a uniform distribution. The center of the plane with coordinates (500,500) was set as the starting position. Afterwards the optimal distance to get all the collectibles was calculated as well as the total distance using the greedy algorithm. This was repeated 1000 times. Note that, the actual choice of plane size does not influence our results as our only interest is the relative performance of the greedy algorithm as compared to the optimal solution.

\begin{figure}[t]
\centering
\includegraphics[scale=.45]{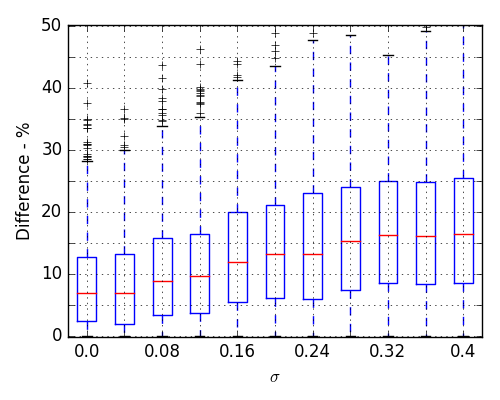}
\caption{Average performance of the greedy algorithm with error compared to the optimal solution, for $N=10$ and different levels of error.}
\label{fig:sigma_diff}
\end{figure}

We found that the performance of the greedy algorithm deteriorates as the number of collectibles $N$ increases (Fig.~\ref{fig:box_plots}). However, the overall performance was surprisingly good and for eleven collectibles $N=11$ it was on average only $7.3\%$ worse than the optimal solution, with upper and lower quartiles at $2.6\%$ and $13.9\%$ respectively.

Next, let us turn our attention to the greedy algorithm with error. This algorithm  is not only worse than the optimal, but it is also worse than the greedy algorithm (Fig.~\ref{fig:starting_location_dependence}). Further analysis was performed in order to estimate what effect the level of error in guestimating the distance has on the performance of the greedy algorithm with error. To this end, the number of collectibles was fixed to $N=10$  and the variance ($\sigma$) in the error term was varied. As expected, higher errors when estimating the distance between two points lead to increasingly poor performance up to $16.9\%$ worse than the optimal~(Fig.~\ref{fig:sigma_diff}).

\section{Conclusion}
\label{conclusion}

In this manuscript we looked at how a greedy algorithm compares to exhaustive search for a variant of the Traveling Salesman Problem, which is relevant in many open world computer games. The main advantages of the greedy algorithm is that it can be performed by a player as he or she is playing, since no complex computations are required, as opposed to exhaustive search which can only be done by a computer. We found that the performance of the greedy algorithm is comparable to the optimal solution, with the difference depending on the number of collectibles and choice of starting position. For $N=11$ the greedy algorithm resulted in routes $7.3\%$ longer on average.

Even when human errors in estimating the distances were included the greedy algorithm performed on average less than $16.9\%$ worse than the optimal solution, depending on the level of error. Our analysis was unfortunately limited by the fact that finding the optimal solution is NP hard with complexity $N!$, which prevented us from calculating the performance for $N > 13$. Nevertheless, the results suggest that players who use the greedy strategy of always going after the closest item first, are in fact playing close to optimally in terms of gathering collectibles.

\end{document}